\documentclass{article}

\usepackage[preprint]{neurips_2026}

\usepackage[utf8]{inputenc} 
\usepackage[T1]{fontenc}    
\usepackage{hyperref}       
\usepackage{url}            
\usepackage{booktabs}       
\usepackage{amsfonts}       
\usepackage{amsmath}  
\usepackage{nicefrac}       
\usepackage{microtype}      
\usepackage{xcolor}         
\usepackage{multirow}
\usepackage{graphicx}

\title{Graph Set Transformer}

\author{%
  Jose E. Escrig Molina \\
  Bioinformatics Group \\
  Wageningen University \\
  Droevendaalsesteeg \\ 
  6708 PB Wageningen \\ 
  The Netherlands \\
  \And
  Baoquan Chen \\
  Department of Physics \\ 
  Technical University of Munich \\
  85748 Garching \\
  Germany \\
  \And
  Daniel~Probst \\
  Bioinformatics Group \\
  Wageningen University \\
  Droevendaalsesteeg \\ 
  6708 PB Wageningen \\ 
  The Netherlands \\
  \texttt{daniel.probst@wur.nl} \\
}

\begin{document}

\maketitle

\begin{abstract}
We introduce the Graph Set Transformer (GST), a neural network architecture for learning on sets of graphs, designed for tasks in which per-element predictions depend on set-wide context as well as local structure. Existing architectures, including DeepSets and SetTransformer, require pre-encoded graph embeddings from a separate GNN, creating a bottleneck between feature extraction and set-level contextualisation. In contrast, GST interleaves node-level feature propagation and cross-graph contextual modelling at every layer, fusing the two levels of information through a gating mechanism. We evaluate GST on a controlled synthetic suite designed to isolate set-conditional structural reasoning and on three real-data benchmarks spanning per-atom reaction-centre identification, reaction yield prediction, and image classification. Under matched parameter budgets, GST performs better than the baselines across these settings. An architectural ablation strongly suggests that the interleaving of local and set context contributes substantially to this advantage.
\end{abstract}

\section{Introduction}
A set is a fundamental mathematical structure and data type that provides a permutation-invariant representation of a collection of unordered elements, lacking an inherent structure such as that of sequences or vectors. This property, while making sets suitable for modelling data without or with unknown sequential structure, has also proven challenging for machine learning \cite{Vinyals2015}. However, recent advances, including DeepSets, Set Transformer, or PointNet, have introduced permutation-invariant deep learning architectures suitable for sets \citet{Zaheer2017,Charles2017,Lee2019}. While DeepSets and Set Transformer require data to be represented as sets of feature vectors, PointNet specialises in point clouds. Recently, \citet{Chinello2025} have introduced the convolutional set transformer (CST), which learns directly on sets of images. 

\begin{figure*}[h]
    \centering
    \includegraphics[width=0.8\textwidth]{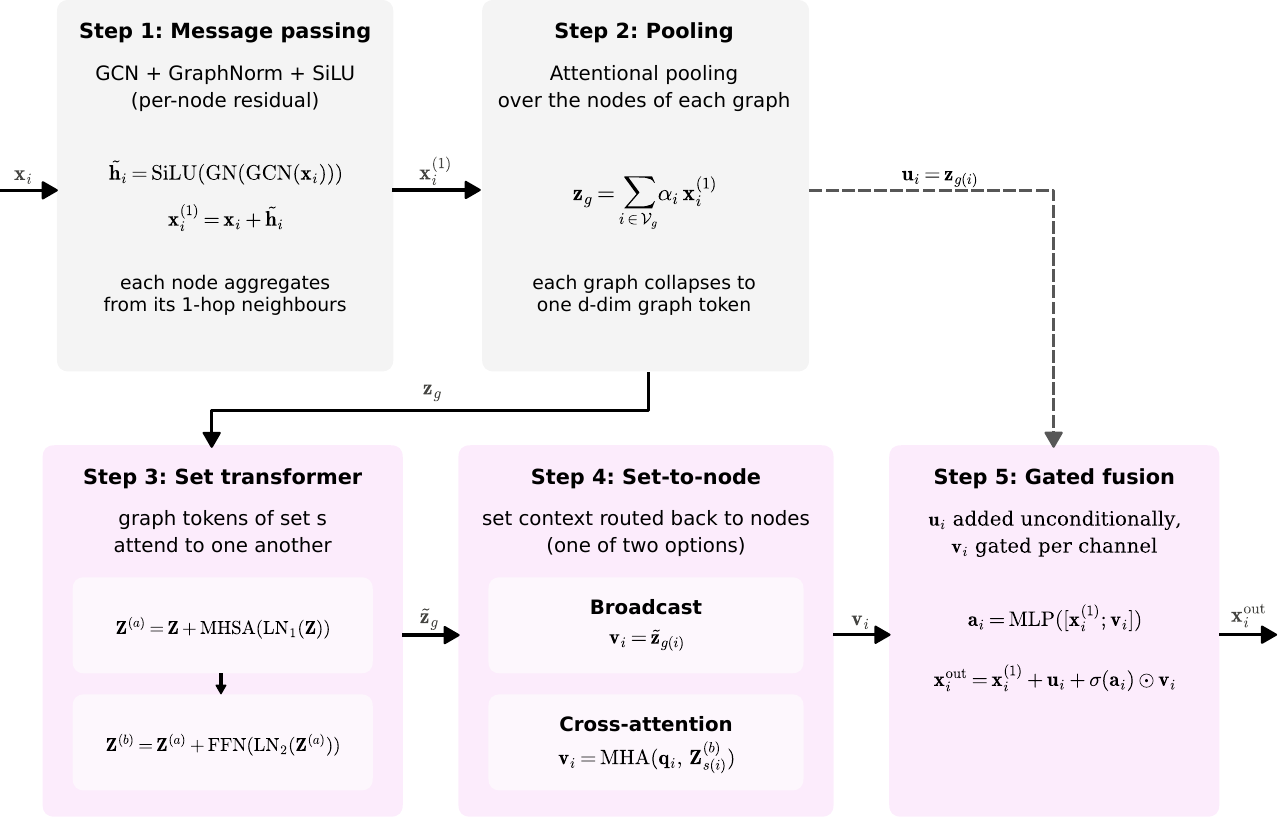}
    \caption{Architecture of the graph set transformer block.}
    \label{fig:schema}
\end{figure*}

Here, we present a generalisation of their architecture to graphs, the graph set transformer (GST). Unlike DeepSets or Set Transformer, which take fixed vector embeddings of each graph as input, GST learns directly on graphs by injecting set-level information back into node features at every layer of a graph neural network that accepts sets of graphs as input (Figure~\ref{fig:schema}). The contrast of our proposed architecture to the standard pipeline, which is to encode graphs independently and then combine the resulting embeddings with a permutation-invariant set head, is architectural rather than incremental. GST interleaves feature extraction and set contextualisation, while pipelines stage them sequentially. We argue that this interleaving architecture removes a bottleneck between graph encoding and set reasoning, where set-level information has to be reconstructed from a single late-stage summary against per-graph features that were computed without any awareness of the rest of the set. Across a controlled synthetic benchmark designed to probe set-conditional structural reasoning and a chemical reaction yield prediction benchmark, the proposed architecture performs better than parameter-matched baselines (GCN followed by DeepSets, GCN followed by SetTransformer) across settings. Code associated with this study can be found here: \url{https://github.com/daenuprobst/gst-conference}.

\section{Related work}
\subsection{DeepSets}
DeepSets, introduced by \citet{Zaheer2017}, defines a deep network architecture for learning permutation-invariant functions over sets. It is based on the theorem stating that a function $f(X)$ operating on a set $X$ is a valid set function, and invariant to the permutation of instances in $X$, iff it can be decomposed in the form $\rho\left(\sum_{x\in X}\phi(x)\right)$, where the summation can be replaced with any pooling function. Functions $\phi$ and $\rho$ are parametrised by neural networks. While the pooling-based aggregation is computationally efficient, each element is encoded independently by $\rho$. This limits the representational power of the architecture as potentially relevant information about inter-element relationships is discarded. This limitation is addressed by the Set Transformer.

\subsection{Set Transformer}
The Set Transformer architecture, introduced by \citet{Lee2019}, addresses the central limitation of DeepSets by implementing an encoder-decoder structure that leverages attention mechanisms to capture inter-element relationships. The encoder consists of induced set attention blocks (ISABs), which apply a sparse Gaussian processes-inspired inducing point method to efficiently compute attention. In the decoder, Lee et al. first aggregate the latents using a pooling by multihead attention (PMA) component, followed by self-attention blocks (SABs), which enable elements in a set to attend to one another.

\subsection{Convolutional set transformer}
The convolutional set transformer (CST) was recently introduced by \citet{Chinello2025} and laid the groundwork for the graph set transformer presented in this article. CST takes visually heterogeneous image sets that share high-level semantics as inputs, a concept that we generalise from Euclidean space to non-Euclidean domains. Conceptually, this requires a shift from the pixel--image--set hierarchy to a node--neighbourhood--graph--set hierarchy, which is introduced by the graph convolutional network. We replace the additive dynamic bias, which uniformly adds a shift across spatial locations, with a more expressive learned gating mechanism that enables node-specific blending of global and local information.

\subsection{Applications}
While the use of graphs is ubiquitous in chemistry, the use of sets has received relatively little attention. Examples include recent work on molecular set representation learning by \citet{boulougouri2024}, who used sets of atoms or molecules to predict molecular properties, reaction yields, and binding affinities, or \citet{Goldman2023}, who used the Set Transformer architecture to learn mass spectra as sets of peaks. Another notable application concerns predicting chemical properties from ensembles, or sets, of molecular conformers \citep{zhu2024learning}. However, within the scope of this study, we chose to avoid the introduction of geometry-aware convolutional layers. By excluding methods that account for molecular geometry, we aim to isolate the representational capacity of set-based architectures and restrict ourselves to the graph convolutional operator introduced by \citet{Kipf2016}.
\section{Methodology}
\subsection{Proposed architecture}

We introduce the graphsetconv block, which processes sets of graphs, where each sample is a set $S$ of graphs
$\mathcal{G}_S = \{G_1, G_2, \ldots\}$ that optimally share a common set context. The complete architecture is a stack of $L$ identical blocks, however, we limit ourselves to the description of a single block here. Let $\mathbf{x}_i \in \mathbb{R}^d$ be the input features of node $i$. The block
produces $\mathbf{x}_i^{\mathrm{out}} \in \mathbb{R}^d$ in six steps. We use  $\mathbf{x}_i^{(1)}$ for the intermediate state. For readability, we omit dropout and stochastic depth, which are applied after each FFN and on each residual branch. We test two variants of the architecture named GST-bc and GST-ca, where the set context is either broadcast (GST-bc) to the nodes or where each node attends via a query to tokens of its parent set (see Step 4 for details).

\paragraph{Step 1} consists of a simple GCN, through which each node aggregates features from its 1-hop neighbours, with GraphNorm providing per-graph normalisation. The result of the normalisation is then added back through a residual. We prefer GraphNorm over LayerNorm as graphs in a set may differ substantially in size. GraphNorm normalizes node activations in a graph independently, thereby keeping graph-level activation magnitudes stable. \citep{touvron2021}.

\begin{equation}
\tilde{\mathbf{h}}_i =
    \mathrm{SiLU}(\mathrm{GN}(\mathrm{GCN}(\mathbf{X})_i; g(i)))
\end{equation}

\begin{equation}
\mathbf{x}_i^{(1)} = \mathbf{x}_i
  + \tilde{\mathbf{h}}_i
\end{equation}
 
\paragraph{Step 2}
summarises each graph in the set into a single $d$-dimensional token $z_g$ through attentional pooling via a softmax-weighted sum of node features $x_i^{(1)}$ whose weights $\alpha_i$ come from a shallow scoring network producing the scalar $e_i$. This forms a bottleneck through which cross-graph information flows. Note that each graph normalises individually by generating its own distribution of $\alpha$.

\begin{equation}
e_i = \mathbf{w}_p^\top\, \mathrm{ReLU}(\mathbf{W}_p\, \mathbf{x}_i^{(1)} + \mathbf{b}_p)
\end{equation}
\begin{equation}
\alpha_i = \frac{\exp e_i}{\sum_{j \in \mathcal{V}_{g(i)}} \exp e_j}
\end{equation}
\begin{equation}
\mathbf{z}_g = \sum_{i \in \mathcal{V}_g} \alpha_i\, \mathbf{x}_i^{(1)}
\end{equation}
 
\paragraph{Step 3} passes information between graphs in a set through per graph tokens that attend one another via an MHSA block. Note that this is the only place in the architecture where information passes between graphs within a set. The transformer learns set-wide context (such as extremum identification, distribution summaries, ranking), which are then written back into the updated token of each graph $\tilde{\mathbf{z}}_g$. For each set of graphs $S$, we stack the per-graph tokens $\mathbf{Z}_s = [\mathbf{z}_g]_{g \in
\mathcal{G}_s} \in \mathbb{R}^{|\mathcal{G}_s| \times d}$ We implement this by padding sets of different cardinality per batch to a common length $K^{\max}$ with mask $\mathbf{M}_s \in \{0,1\}^{K^{\max}}$).

\begin{equation}
\mathbf{Z}_s^{(\mathrm{a})}
  = \mathbf{Z}_s
    + \mathrm{MHSA}(\mathrm{LN}_1(\mathbf{Z}_s),\, \mathbf{M}_s)
\end{equation}

\begin{equation}
\mathbf{Z}_s^{(\mathrm{b})}
  = \mathbf{Z}_s^{(\mathrm{a})}
    + \mathrm{FFN}(\mathrm{LN}_2(\mathbf{Z}_s^{(\mathrm{a})}))
\end{equation}
Let $\tilde{\mathbf{z}}_g$ denote the row of $\mathbf{Z}_{s(g)}^{(\mathrm{b})}$
corresponding to graph $g$ (the post-set-transformer token).
 
\paragraph{Step 4} routes the set context back into individual nodes. $\mathbf{u}_i = \mathbf{z}_{g(i)}$ is the
pre-set-transformer pooled token of the parent graph of node $i$. $\mathbf{v}_i$ comes from the post-set-transformer tokens and is derived either through broadcasting
\begin{equation}
\mathbf{v}_i = \tilde{\mathbf{z}}_{g(i)}.
\end{equation}

or cross-attention, where each node attends via a query to the post-set-transformer tokens of its parent set $|\mathcal{G}_{s(i)}|$:
\begin{equation}
\mathbf{v}_i =
      \mathrm{MHA}(\mathbf{q}_i^\top,\, \mathbf{Z}_{s(i)}^b,\, \mathbf{M}_{s(i)})
\end{equation}
where $\mathbf{q}_i = \mathrm{LN}_q(\mathbf{x}_i^{(2)}) \in \mathbb{R}^d$.

\paragraph{Step 5} fuses the two context vectors $\mathbf{u}_i$ and $\mathbf{v}_i$  into the
residual stream through learned asymmetric gates conditioned on
$[\mathbf{x}_i^{(1)};\, \mathbf{v}_i]$. Each node thereby decides how much of each context to absorb per channel. I.e., a node whose home graph contains the necessary information can ignore set context. The asymmetry reflects an inductive bias where the parent graph of a node is always deemed important and the influence of the set is decided based on local context. A shallow MLP creates two $d$-dimensional gates
$[\mathbf{x}_i^{(2)};\, \mathbf{v}_i]$:

\begin{equation}
[\mathbf{a}_i;\, \mathbf{b}_i]
  = \mathbf{W}_{g2}\,\mathrm{SiLU}\!\big(\mathbf{W}_{g1}
       [\mathbf{x}_i^{(2)};\, \mathbf{u}_i;\, \mathbf{v}_i] + \mathbf{b}_{g1}\big)
     + \mathbf{b}_{g2}
  \;\in\; \mathbb{R}^{2d}
\end{equation}

\begin{equation}
\mathbf{x}_i^{\mathrm{out}}
  = \mathbf{x}_i^{(2)}
   + \sigma(\mathbf{a}_i) \odot \mathbf{u}_i
   + \sigma(\mathbf{b}_i) \odot \mathbf{v}_i,
\end{equation}

where $\mathbf{W}_{g1} \in \mathbb{R}^{d \times 3d}$,
$\mathbf{W}_{g2} \in \mathbb{R}^{2d \times d}$, and $\mathbf{b}_{g2}$ are initialized to $\mathbf{0}$. The full model is a stack of $L$ such blocks without parameter sharing, followed by an MLP head producing the task output:
\begin{equation}
    \mathbf{x}_i^{[L]} = \mathrm{GST}^{[L]} \circ \cdots \circ \mathrm{GST}^{[1]}(\mathbf{x}_i^{[0]}),
    \qquad
    y_i = \mathrm{MLP}_{\mathrm{head}}(\mathbf{x}_i^{[L]}).
\end{equation}

\subsubsection{Experimental design}
For all experiments, we compare GST with three layers (using \texttt{GCNConv}) as a basis to DeepSets and SetTransformer, both applied to graph embeddings produced by a graph neural network with three graph convolutional layers (\texttt{GCNConv}). The hyperparameters of all models were tuned on the synthetic data. The number of hidden dimensions of all graph convolutional layers in all models were set at 128. For experiments with parameter parity, we reduced the hidden dimensions of the GST model. All experiments are set-to-global tasks, where we train on set-level labels and predict set-level labels. For the synthetic benchmark, we reduced the number of layers of all models to two. All models were trained and evaluated on a single RTX 4070TI and RTX 5090 GPUs.

\paragraph{Synthetic benchmark} For the initial evaluation of the proposed architecture, we designed a benchmark on synthetic data. We train our architecture as well as the DeepSets and SetTransformer baselines on sets of randomly generated Erdős–Rényi graphs ($n \in \{12, \ldots, 25\}$, $p \in [0.10, 0.25]$, $|S| \in \{3, \ldots, 7\}$). The sets are generated at train, validation, and test time, resulting in 40,000 training, 640 validation, and 512 test sets. All nodes $v$ in a set $S$ with the maximum degree per set are declared anchor nodes:
\begin{equation}
     \mathcal{A}(S) = \{v \in V(S) : \deg(v) = d^\star(S)\} \quad\text{where} \quad d^\star(S) = \max_{v \in V(S)} \deg(v)
\end{equation}
For the RING-K experiment, every node $v$ then receives a binary label $y(v)$, where $y(v)=1$ iff the shortest-path distance from $v$ to the nearest anchor within its parent graph is exactly $K$, and $y(v)=0$ otherwise. The task of the model is the binary classification of the nodes. Similarly, in the BALL-K experiments, each node receives the label $y(v)=1$ iff $\text{dist}(v,A(S)) \leq K$ and $y(v)=0$ otherwise. For DIST, the task is to predict the shortest distance to an anchor, where the nodes are labelled by $y(v) = \min\big(\text{dist}(v, \mathcal{A}(S)),\, D_{\max}\big)$. Finally, EXTRAPOLATE is RING-K with $K=2$ where the set cardinalities $|S|$ differ between train/validation and test sets. Training and validation sets have cardinality $|S|\in\{3,4,5\}$ while test sets are evaluated at $|S|=8$, $|S|=10$, and  $|S|=12$ in separate runs. The goal of this last task is to evaluate the models under shifting distributions of anchorless graphs and set-wide max degrees. For RING-K, BALL-K, and EXTRAPOLATE we report average precision (AP), for DIST we report MAE. For this experiment, we reduced the hidden layer sizes of GST-bc and GST-ca to 28 and 32, respectively, to match the parameter count of GCN-ds (39,175 for GST-bc, 42,019 for GST-ca, 38,081 for GCN-ds, and 71,745 for GCN-st). The baselines were both kept at a hidden dimension of 64.

\paragraph{CIFAR-10} Initially, we evaluate the performance of GST on the CIFAR-10 ($n=60,000$) data set \citep{krizhevsky2009learning}. We chose an image multi-class classification task to assess whether the image-based performance characteristics measured by \citet{Chinello2025} compare to those of our graph-based approach. However, we do not use raw image data but graph-encoded images instead and therefore refrain from direct comparisons of accuracy magnitude \citep{Dwivedi2023}. On the CIFAR-10 data set, the models were evaluated on five different set cardinalities (5, and 10). . All results are averaged over 3 runs with different random seeds, and we report the mean and standard deviation. We report the macro-averaged ROC-AUC and PR-AUC, computed using a one-vs-rest strategy.

\paragraph{Reaction yield prediction} We predict the yield of Buchwald-Hartwig reactions by creating sets of molecular graphs from reaction participants. Each reaction is therefore represented as a set $S$. We use the data provided by \citet{Ahneman2018} with common splits. Specifically, we run a training-set-size ablation where the test fraction is being held constant at 30\% while sweeping the training fraction across seven settings from 0.025 to 0.700. We use 10\% of the training set as a validation set (capped at $N_\text{val}=50$) for early stopping. We report RMSE, MAE, and $\text{R}^2$. For this experiment, we reduced the hidden layer sizes of GST-bc and GST-ca to 48 and 44, respectively, to match the parameter count of GCN-ds (139,684 for GST-bc, 142,256 for GST-ca, 137,729 for GCN-ds, and 270,593 for GCN-st). The baselines were both kept at a hidden dimension of 128.

\paragraph{Reaction centre prediction} We further evaluate GST on the USPTO-15K reaction centre identification benchmark by ~\citet{uspto15k}, which is formulated as per-atom binary classification reactions. As in reaction yield prediction, each sample is the set of molecular graphs (the reaction participants) and the per-atom label $y(v)=1$ iff that atom is part of a reaction centre, which is defined as the union of atoms whose hydrogen counts change and atoms incident to bonds that are broken or formed in going to the product, and $y(v)=0$ otherwise. Reaction-centre atoms are sparse at approximately 5\% of all atoms, and crucially, the label is set-conditional. Whether a given atom reacts depends not only on its parent molecule, but also on which other molecules participate in the reaction. This makes the task a chemical application of the synthetic RING-K diagnostic. For this experiment, we report parameter counts in Table \ref{tab:uspto15k}.

\subsubsection{Implementation details}
All models are implemented in PyTorch~\cite{Paszke2019} using PyTorch Geometric~\cite{Fey2019}. We use 3 graph set convolutional layers (\texttt{GCNConv}) with a hidden dimension of 64. For the Set Transformers model, the multi-head attention heads were set to 4, and the number of inducing points was set to 32. The DeepSets model does not use any attention mechanism. For the Set Graph Transformer model, the number of multi-head attention layers was also set to 4. A base dropout with probability 0.1 is applied after each layer to prevent overfitting. We train using the AdamW optimiser with a learning rate of $1\times 10^{-3}$ for DeepSets and Set Transformer and $1\times 10^{-4}$ for GST (lowering the learning rate for DeepSets and Set Transformer was not beneficial to their performance)~\citep{loshchilov2019decoupled}. All results are gathered using early stopping on the validation metric. For the CIFAR-10 dataset experiments, the Adam optimiser was used, and all the learning rates were set to $1\times 10^{-3}$.

\section{Results and discussion}
\subsection{Synthetic experiments}
Our synthetic experiments validate the design rationale of the introduced architecture (Table \ref{tab:synth_main}). In the RING-K experiments, the gap between GST and the baselines grows with increasing $K$ by approximately 5, 9, and 15 AP, /respectively. Predicting RING-K labels require composing three operations: (1) extracting a set-wide statistic, (2) marking the anchors, (3) and then propagate distance information through edges. While all architectures perform substantially worse at $K=3$ due to the shared depth bottleneck of two layers, the DeepSet and SetTransformer-based architectures additionally degrade with increasing $K$ due to their message-passing layers computing only local embeddings that are not combined with the set summary until the final head.

\begin{table}[h]
\centering
\small
\setlength{\tabcolsep}{4pt}
\caption{Synthetic-benchmark results.\ GST variants GST-bc and GST-ca vs.\ GCN+DS and GCN-ST pipelines over 5 seeds. \textbf{bold} denotes best model in row.}
\label{tab:synth_main}
\resizebox{0.8\textwidth}{!}{%
\begin{tabular}{lcccc}
\toprule
Setting & GST-ca & GST-bc & GCN-DS & GCN-ST \\
\midrule
\multicolumn{5}{l}{\textbf{RING-K}\ (AP\ $\uparrow$)} \\
$K=1$ & 0.999\,$\pm$\,0.001 & \textbf{1.000\,$\pm$\,0.000} & 0.924\,$\pm$\,0.011 & 0.948\,$\pm$\,0.007 \\
$K=2$ & 0.965\,$\pm$\,0.006 & \textbf{0.972\,$\pm$\,0.005} & 0.846\,$\pm$\,0.019 & 0.884\,$\pm$\,0.010 \\
$K=3$ & 0.483\,$\pm$\,0.033 & \textbf{0.520\,$\pm$\,0.029} & 0.341\,$\pm$\,0.025 & 0.371\,$\pm$\,0.025 \\
\addlinespace
\multicolumn{5}{l}{\textbf{BALL-K}\ (AP\ $\uparrow$)} \\
$K=0$ & 0.967\,$\pm$\,0.043 & \textbf{0.984\,$\pm$\,0.019} & 0.827\,$\pm$\,0.020 & 0.881\,$\pm$\,0.019 \\
$K=1$ & 1.000\,$\pm$\,0.000 & \textbf{1.000\,$\pm$\,0.000} & 0.932\,$\pm$\,0.011 & 0.957\,$\pm$\,0.007 \\
$K=2$ & \textbf{0.991\,$\pm$\,0.001} & 0.984\,$\pm$\,0.014 & 0.927\,$\pm$\,0.016 & 0.944\,$\pm$\,0.014 \\
\addlinespace
\multicolumn{5}{l}{\textbf{DIST}\ (MAE\ $\downarrow$)} \\
$K=5$ & 0.181\,$\pm$\,0.043 & \textbf{0.142\,$\pm$\,0.043} & 0.524\,$\pm$\,0.027 & 0.441\,$\pm$\,0.026 \\
\addlinespace
\multicolumn{5}{l}{\textbf{EXTRAPOLATE}\ (AP\ $\uparrow$)} \\
$K=2$, $|S_\text{test}|=8$ & 0.970\,$\pm$\,0.005 & \textbf{0.976\,$\pm$\,0.005} & 0.826\,$\pm$\,0.010 & 0.871\,$\pm$\,0.018 \\
$K=2$, $|S_\text{test}|=10$ & 0.970\,$\pm$\,0.007 & \textbf{0.977\,$\pm$\,0.005} & 0.809\,$\pm$\,0.010 & 0.868\,$\pm$\,0.021 \\
$K=2$, $|S_\text{test}|=12$ & 0.969\,$\pm$\,0.008 & \textbf{0.974\,$\pm$\,0.009} & 0.786\,$\pm$\,0.014 & 0.858\,$\pm$\,0.022 \\
\addlinespace
\midrule
\textit{Parameters} & 37,795 & 36,759 & 38,081 & 71,745 \\
\bottomrule
\end{tabular}
}
\end{table}

BALL-K helps us decompose the observed deficit of the two baselines in RING-K into a set-extraction component and an interleaving component. The lack of a substantially growing gap for $K=1$ and $K=2$ in BALL-K identifies the architecture (the interleaving component) of GST as the source of the growing gap on RING-K under exact-distance requirements. This is due to its ability to compose set-conditional anchor identification with hop-by-hop distance propagation at every layer, compared to only after message passing has terminated. The gap between GST and the baselines for $K=0$ highlights their deficit in the set-extraction component. With $K=0$, where only the anchor receives a 1 label and local-degree heuristics substitute for set-wide reasoning. This is similar to the drop of the baselines between $K=1$ and $K=2$ in RING-K. With the DIST experiment, where the models have to regress exact hop counts, we show that the gap between GST and baselines survives the removal of any threshold-based slack from the classification experiments. Finally, under set-size shift, the performance of GST remains stable and high while the baselines degrade monotonically. This is evidence that the interleaved architecture of GST has captured the set operation rather than simply learned training-set-size statistics.

\subsection{Chemistry Applications}

\paragraph{Reaction yield prediction} Both GST variants perform better than the baselines across training set fraction size, with the largest gains in the noisy low-data regime at $n_{\text{train}} = 90$ (2.5\%), where GST-bc $R^2 = 0.53$ and GCN-ST $R^2 = 0.18$ (Table \ref{tab:results}). The gap between the architectures narrows but persists with increasing fraction size. At 70\% of the data, GST-ca reaches $R^2 = 0.95$ compared to $0.89$ for the strongest baseline. The two GST variants are approximately tied at every fraction, reproducing our findings in the synthetic benchmark, where we showed the interleaving inductive bias leads to an architectural advantage.

\begin{table}[h]
\centering
\caption{Performance comparison across train fractions (mean $\pm$ std, from 5 runs). \textbf{Bold} denotes best $R^2$ in fraction.}
\label{tab:results}
\resizebox{0.8\textwidth}{!}{%
\begin{tabular}{llccc}
\toprule
Train frac. ($n_{\text{train}}$) & Model & RMSE $\downarrow$ & MAE $\downarrow$ & $R^2$ $\uparrow$ \\
\midrule
\multirow{4}{*}{\textbf{0.025} (90)}
 & GST-bc            & 18.716 $\pm$ 2.090 & 14.433 $\pm$ 2.317 & \textbf{0.531 $\pm$ 0.107} \\
 & GST-ca  & 20.147 $\pm$ 3.863 & 15.182 $\pm$ 3.191 & 0.450 $\pm$ 0.227 \\
 & GCN+DS               & 20.372 $\pm$ 2.566 & 15.250 $\pm$ 2.109 & 0.444 $\pm$ 0.141 \\
 & GCN-ST         & 24.732 $\pm$ 3.173 & 18.509 $\pm$ 3.711 & 0.180 $\pm$ 0.212 \\
\midrule
\multirow{4}{*}{\textbf{0.050} (179)}
 & GST-bc            & 16.482 $\pm$ 2.796 & 11.798 $\pm$ 2.529 & 0.633 $\pm$ 0.133 \\
 & GST-ca  & 16.189 $\pm$ 1.541 & 11.294 $\pm$ 1.183 & \textbf{0.650 $\pm$ 0.067} \\
 & GCN+DS               & 19.886 $\pm$ 6.509 & 15.336 $\pm$ 6.195 & 0.445 $\pm$ 0.364 \\
 & GCN-ST         & 19.644 $\pm$ 2.332 & 14.031 $\pm$ 2.029 & 0.483 $\pm$ 0.120 \\
\midrule
\multirow{4}{*}{\textbf{0.100 }(357)}
 & GST-bc            & 12.505 $\pm$ 1.204 & 8.657 $\pm$ 1.211  & \textbf{0.791 $\pm$ 0.042} \\
 & GST-ca  & 13.273 $\pm$ 0.899 & 9.307 $\pm$ 0.762  & 0.765 $\pm$ 0.032 \\
 & GCN+DS               & 13.926 $\pm$ 1.580 & 9.824 $\pm$ 1.201  & 0.741 $\pm$ 0.060 \\
 & GCN-ST         & 15.755 $\pm$ 2.561 & 11.309 $\pm$ 1.982 & 0.666 $\pm$ 0.113 \\
\midrule
\multirow{4}{*}{\textbf{0.200} (741)}
 & GST-bc            & 10.486 $\pm$ 0.708 & 6.839 $\pm$ 0.552 & 0.854 $\pm$ 0.020 \\
 & GST-ca  & 10.177 $\pm$ 0.482 & 6.783 $\pm$ 0.459 & \textbf{0.862 $\pm$ 0.013} \\
 & GCN+DS               & 12.593 $\pm$ 1.881 & 8.913 $\pm$ 1.372 & 0.787 $\pm$ 0.064 \\
 & GCN-ST         & 12.630 $\pm$ 1.358 & 8.719 $\pm$ 1.110 & 0.787 $\pm$ 0.046 \\
\midrule
\multirow{4}{*}{\textbf{0.300} (1136)}
 & GST-bc            & 9.019 $\pm$ 1.890  & 5.773 $\pm$ 1.335 & \textbf{0.889 $\pm$ 0.049} \\
 & GST-ca  & 9.753 $\pm$ 2.296  & 6.324 $\pm$ 2.040 & 0.870 $\pm$ 0.065 \\
 & GCN+DS               & 11.204 $\pm$ 1.659 & 7.592 $\pm$ 1.462 & 0.831 $\pm$ 0.053 \\
 & GCN-ST         & 12.793 $\pm$ 1.187 & 8.642 $\pm$ 1.069 & 0.781 $\pm$ 0.039 \\
\midrule
\multirow{4}{*}{\textbf{0.500} (1,928)}
 & GST-bc            & 7.348 $\pm$ 0.723  & 4.667 $\pm$ 0.478 & \textbf{0.928 $\pm$ 0.014} \\
 & GST-ca  & 8.009 $\pm$ 0.726  & 5.088 $\pm$ 0.580 & 0.914 $\pm$ 0.015 \\
 & GCN+DS               & 10.205 $\pm$ 3.670 & 6.973 $\pm$ 3.235 & 0.852 $\pm$ 0.115 \\
 & GCN-ST         & 10.278 $\pm$ 1.819 & 6.651 $\pm$ 1.389 & 0.857 $\pm$ 0.048 \\
\midrule
\multirow{4}{*}{\textbf{0.700} (2,718)}
 & GST-bc            & 6.392 $\pm$ 0.756 & 4.162 $\pm$ 0.367 & 0.945 $\pm$ 0.013 \\
 & GST-ca  & 6.379 $\pm$ 0.392 & 4.029 $\pm$ 0.261 & \textbf{0.946 $\pm$ 0.007} \\
 & GCN+DS               & 8.379 $\pm$ 1.840 & 5.472 $\pm$ 1.196 & 0.904 $\pm$ 0.044 \\
 & GCN-ST         & 9.056 $\pm$ 1.921 & 5.810 $\pm$ 1.324 & 0.888 $\pm$ 0.051 \\
\bottomrule
\end{tabular}
}
\end{table}

Given this result, we theorise that the interleaving inductive bias means that atom-level features of a molecule are updated within the context of the other molecules in the reaction at every layer, opposed to each molecule being encoded in isolation before being passed to a set encoder. This makes sense from a chemistry perspective, as the reactive behaviour of an atom depends on all reaction participants, not just its own molecule. Encoding each molecule in isolation deprives atoms of this reaction-wide context.

\paragraph{Reaction centre prediction}
Similar to RING-K, USPTO-15K reaction-centre prediction is a per-node classification task where the correct label for each atom depends on set-wide context. As can be seen in Table \ref{tab:uspto15k}, the results of the synthetic benchmark hold for the real-world data set. GST-bc performs best across changes in number of layers and parameters, with GST-ca a close second and both baselines trailing. At natural budgets and a single layer (block 3), GST-bc reaches $0.879$~AP versus $0.822$ for GCN-ds and $0.821$ for GCN-st, a gap of $5.7$~AP. Adding depth (blocks 4 and 5) narrows the gap but does not close it. This can be seen as the inverse of increasing $K$ in RING-K.

\begin{table}[h]
\centering
\small
\caption{USPTO-15K reaction-centre prediction (per-atom AP, mean $\pm$ std, from 5 runs).\textbf{Bold} denotes the best AP within each block.}
\label{tab:uspto15k}
\begin{tabular}{llrrr}
\toprule
Depth & Model & AP $\uparrow$ & Parameters & Hidden \\
\midrule
\multirow{4}{*}{1 layer}
 & GST-ca & 0.839\,$\pm$\,0.004 & 130,950 & 76 \\
 & GST-bc & \textbf{0.858\,$\pm$\,0.005} & 130,790 & 84 \\
 & GCN-ds           & 0.822\,$\pm$\,0.001 & 134,913 & 128 \\
 & GCN-st     & 0.819\,$\pm$\,0.003 & 267,777 & 128 \\
\midrule
\multirow{4}{*}{1 layer}
 & GST-ca & 0.868\,$\pm$\,0.004 & 366,978 & 128 \\
 & GST-bc & \textbf{0.876\,$\pm$\,0.002} & 300,674 & 128 \\
 & GCN-ds           & 0.848\,$\pm$\,0.002 & 365,913 & 212 \\
 & GCN-st     & 0.824\,$\pm$\,0.004 & 376,353 & 152 \\
\midrule
\multirow{4}{*}{1 layer}
 & GST-ca & 0.869\,$\pm$\,0.003 & 366,978 & 128 \\
 & GST-bc & \textbf{0.879\,$\pm$\,0.002} & 300,674 & 128 \\
 & GCN-ds           & 0.822\,$\pm$\,0.002 & 134,913 & 128 \\
 & GCN-st     & 0.821\,$\pm$\,0.004 & 267,777 & 128 \\
\midrule
\multirow{4}{*}{2 layers}
 & GST-ca & 0.902\,$\pm$\,0.003 & 714,755 & 128 \\
 & GST-bc & \textbf{0.904\,$\pm$\,0.002} & 582,147 & 128 \\
 & GCN-ds           & 0.858\,$\pm$\,0.005 & 151,809 & 128 \\
 & GCN-st     & 0.853\,$\pm$\,0.004 & 284,673 & 128 \\
\midrule
\multirow{4}{*}{3 layers}
 & GST-ca & 0.907\,$\pm$\,0.002 & 1,062,532 & 128 \\
 & GST-bc & \textbf{0.912\,$\pm$\,0.003} & 863,620 & 128 \\
 & GCN-ds           & 0.863\,$\pm$\,0.003 & 168,705 & 128 \\
 & GCN-st     & 0.863\,$\pm$\,0.005 & 301,569 & 128 \\
\bottomrule
\end{tabular}
\end{table}

The results shown in the parameter-equalized blocks one and two hint at overall capacity not being the explanation. When GST is shrunk to baseline-sized parameter counts ($\approx$130K) by reducing its hidden to $76$ and $84$ while baselines retain theirs at 128, GST-bc still performs better by $3.6$~AP. When the baselines are instead grown to the natural parameter count of GST ($\approx$300--376K), GST-bc still wins by $2.8$~AP.

\subsection{Image Classification}
CIFAR-10 Set Classification To evaluate our architecture on high-dimensional vision data, we benchmarked the models on a CIFAR-10 set classification task. As shown in Table \ref{tab:CIFAR-10}, the GST variants consistently outperform the GCN-based baselines across both set sizes. GST-ca achieves the highest performance, peaking at an ROC-AUC of 99.4 and a PR-AUC of 95.2 for sets of size 10. The performance gap is particularly evident in the more sensitive PR-AUC metric. At a set size of 5, GST-ca outpaces the strongest baseline (GCN-ds) by 13.4 points, and GST-bc leads by 6.1 points.

When the set size is increased from 5 to 10, all models show improved performance, indicating that a larger number of images provides more context for the set-level prediction. However, even with this increased context, the baselines fail to reduce the performance gap, with GCN-ds reaching a PR-AUC of 88.3 compared to GST-ca’s 95.2.

These results further corroborate the findings from our synthetic and chemistry experiments. The deficit observed in the baselines is likely due to their disjointed processing, where each image is encoded in isolation before being aggregated into a set representation. In contrast, GST allows the features of individual images to be updated dynamically using the context of the other images in the set at every layer. This suggests that allowing elements to contextualize one another during the extraction phase, rather than just at the final pooling stage, is highly advantageous.

\begin{table}[h]
\centering
\small
\caption{CIFAR-10 image classification (by set), mean $\pm$ std, from 3 runs. \textbf{Bold} denotes the best ROC-AUC and PR-AUC.}
\label{tab:CIFAR-10}
\begin{tabular}{llrrr}
\toprule
Set Size & Model & ROC-AUC & PR-AUC\\
\midrule
\multirow{4}{*}{Set size 5}
 & GST-ca & \textbf{98.3\,$\pm$\,0.1} & \textbf{89.0\,$\pm$\,0.8} \\
 & GST-bc & 96.9\,$\pm$\,0.1 & 81.7\,$\pm$\,0.4 \\
 & GCN-ds           & 95,7\,$\pm$\,0.2 & 75.6\,$\pm$\,0.4 \\
 & GCN-st     & 95,3\,$\pm$\,0.8 & 74.8\,$\pm$\,3.1 \\
\midrule
\multirow{4}{*}{Set size 10}
 & GST-ca & \textbf{99.4\,$\pm$\,0.1} & \textbf{95.2\,$\pm$\,0.4} \\
 & GST-bc & 99.1\,$\pm$\,0.1 & 93.3\,$\pm$\,0.4 \\
 & GCN-ds           & 98.3\,$\pm$\,0.1 & 88.3\,$\pm$\,0.6 \\
 & GCN-st    & 97.9\,$\pm$\,0.4 & 86.7\,$\pm$\,2.1 \\
\bottomrule
\end{tabular}
\end{table}

\subsection{Computational cost}
We evaluate the computational cost empirically on batches of $32$ sets of $4$ independently sampled Erd\H{o}s--R\'enyi graphs with $n \sim \mathcal{U}\{12, \dots, 25\}$, $p \sim \mathcal{U}[0.10, 0.25]$, and node features $x_v \sim \mathcal{N}(0, I_{64})$. The results are shown in Table \ref{tab:timing}.

\begin{table}[h]
\centering
\small
\caption{Wall-clock cost per training step (mean $\pm$ std over [N] steps; median; relative to the fastest model). Measured on [hardware], batch size [B], hidden dimension [d], [L] layer(s).}
\label{tab:timing}
\begin{tabular}{lrrrr}
\toprule
Model & ms/step & median & relative & parameters \\
\midrule
GST-bc  & 38.03\,$\pm$\,1.85 & 37.89 & 4.26$\times$ & 672{,}004 \\
GST-ca & 56.09\,$\pm$\,1.68 & 56.25 & 6.28$\times$ & 870{,}916 \\
GCN-ds           &  8.93\,$\pm$\,0.54 &  8.82 & 1.00$\times$ & 141{,}313 \\
GCN-st      & 11.10\,$\pm$\,0.25 & 11.11 & 1.24$\times$ & 274{,}177 \\
\bottomrule
\end{tabular}
\end{table}

\section{Conclustion and Limitations}

We introduced the Graph Set Transformer (GST), a graph neural architecture for set-of-graphs tasks, which interleaves local message passing with set-level attention at every layer rather than staging the two sequentially. We showed across a controlled synthetic suite designed to isolate set-conditional structural reasoning, USPTO-15K reaction-centre prediction, and Buchwald--Hartwig yield prediction, as well as CIFAIR-10 classification, that GST consistently outperforms parameter-matched DeepSets and SetTransformer baslines, with the largest gains where set context is most informative. We carried out parameter equalization, set-size extrapolation, and the BALL-K decomposition, which together indicate that the advantage stems from the interleaved architecture rather than from raw capacity or task-specific tuning. We recommend the broadcast variant (GST-bc) as the default, as it matches or exceeds the cross-attention variant in accuracy at lower wall-clock and parameter cost.

\paragraph{Limitations} A practical limitation of GST is wall-clock cost. Interleaving set-level attention at every layer multiplies its per-batch runtime compared to the baseline architectures, which compute set context only once. The broadcast variant reduces this overhead with no measurable accuracy loss across our experiments, and is what we recommend as the default when applying GST.


\bibliographystyle{plainnat}  
\bibliography{references}

\newpage

\end{document}